# More than Word Frequencies: Authorship Attribution via Natural Frequency Zoned Word Distribution Analysis

Zhili Chen, Liusheng Huang, Wei Yang, Peng Meng, and Haibo Miao

*Abstract* - **With such increasing popularity and availability of digital text data, authorships of digital texts can not be taken for granted due to the ease of copying and parsing. This paper presents a new text style analysis called natural frequency zoned word distribution analysis (NFZ-WDA), and then a basic authorship attribution scheme and an open authorship attribution scheme for digital texts based on the analysis. NFZ-WDA is based on the observation that all authors leave distinct intrinsic word usage traces on texts written by them and these intrinsic styles can be identified and employed to analyze the authorship. The intrinsic word usage styles can be estimated through the analysis of word distribution within a text, which is more than normal word frequency analysis and can be expressed as: which groups of words are used in the text; how frequently does each group of words occur; how are the occurrences of each group of words distributed in the text. Next, the basic authorship attribution scheme and the open authorship attribution scheme provide solutions for both closed and open authorship attribution problems. Through analysis and extensive experimental studies, this paper demonstrates the efficiency of the proposed method for authorship attribution.**

## I. INTRODUCTION

Authorship attribution examines a text of unknown or disputed authorship and tries to attribute the text to a known author based on texts of known authorships [1]. The main idea is that texts written by different authors are of distinct text styles and by measuring these styles we can distinguish between the texts. Thus, authorship attribution investigates style differences between texts written by different people.

Most of previous researches on authorship attribution were completely or partially based on frequency analysis of most frequent words, especially those also called function words. Function words were found to be among the best features for authorship attribution [2]. Some other researches have applied context words to authorship attribution. They have shown that context words can also contribute to style identification [3][4]. Beside authorship attribution methods using word level characteristics, there are some further aspects of natural language explored, such as syntactic features, semantic features and application-specific features [2]. These features are found normally less effective than word level features for authorship attribution, but can augment to some extent the attribution results when used together with word level features. As a result, word level features play a crucial role in the solution of authorship attribution problems.

In this paper, we examine all words in a given text for its author's writing style. We group all words to some word groups (natural frequency zones) according to their general frequencies in natural language (natural frequencies), instead of using the words themselves directly. We argue that all words in a given text would contribute to analyzing its author's writing style no matter whether the words are function words or context words. The reasons are as follows. First, it is a good consensus that function words contribute a lot to author style identification; second, some researches have initially shown that content words has effect on author style identification [3][4]; third, as described above, in our examination we focus only on the natural frequency property for all words instead of themselves, reducing the effect of text contents as much as possible. A more intuitive argument is that, to express the same meanings, one author may tend to use very frequent content words, while another may tend to employ less frequent, jargon words, causing distribution differences of words of different natural frequencies.

Furthermore, the "word distribution" in our paper comprises three levels (As shown in Fig. 1): which groups of words are used in the text (Vocabulary); how frequently does each group of words occur (Frequency); how are the occurrences of each group of words distributed in the text

(Occurrence). The first two levels of information are also explored similarly in the previous work that based on "bag-of-words". However, the third level of information has been first introduced in our conference paper [5] as far as we know. Different from normal word frequency analysis focusing only on the information of word frequencies, in the third level we further examine the differences of distributions of word occurrences in the text. In Fig. 1, we can see that level vocabulary provides subset of information provided by level frequency, which in turn provides subset of information provided by level occurrence. Additionally, the figure shows that two texts possibly result in the same information (or similar information) of level vocabulary and frequency, but the styles of them can be very different. In this case, it is necessary to take the information of level occurrence into account for style identification. In this paper, we examine carefully the level occurrence to provide more information for authorship attribution of digital texts.

From the descriptions above, we can see that our new method is different from the previous work in authorship attribution in some points. First of all, all words are used for analyzing the writing style. Second, words are identified and grouped only according to their natural frequencies, but not senses, spelling or part of speech, and so on, making the analysis being less content-sensitive intuitively. Third, word occurrence information is retained to provide more information about styles of texts, while the previous method normally uses only the word frequency information.

|  | Vocabulary | Frequency | Occurrence |
|---|---|---|---|
| Text-1 ABAABBABBAABAA | A B | A: 8 B: 6 | A: 0,2,3,6,9,10,12,13 B: 1,4,5,7,8,11 |
| Text-2 AABBBBABBAAAAA | A B | A: 8 B: 6 | A: 0,1,6,9,10,11,12,13 B: 2,3,4,5,7, 8 |

Fig. 1. Three Levels of Word Distribution

Text-1 and Text-2 contain the same two word groups A and B, and they have the same level vocabulary and frequency information. The difference of their styles can only be reflected by the level occurrence information.

The remainder of the paper is organized as follows: In Section II, the detailed style analytic method is presented; next, two authorship attribution schemes are proposed in Section III; then, experiments and analysis are carried out in Section IV; related work is given in Section V; finally, section VI is the conclusions and future work.

## II. STYLE ANALYTIC METHOD

As already described in Section I, the proposed text style analytic method called natural frequency zoned word distribution analysis (NFZ-WDA) will explore the occurrence information of all words in texts. Besides, previous researches have shown that a wealth of features, many of which may be weak discriminators, almost always offer more tenable results than a smaller number of strong ones [6]. Therefore, in our analysis, we group words into a lot of natural frequency zones (NFZs) according to their natural frequencies, and for each NFZ, design two features describing the distributions of NFZ word occurrences, resulting in a great number of features. We use the large number of features to represent the writing styles of authors.

The original version of the analysis was firstly introduced in our conference paper [5]. However, in the current version we have designed two more ways of NFZ partition, corrected some errors, added some more explanations and examples, and rearranged the presentation more clearly. In this presentation, the analysis is depicted in four steps as follows.

### A. Word Occurrence Computation

In this step, occurrence information of all words in the given text is represented and computed for later use.

As done in papers [5][7], in the analysis a given text is simply viewed as a sequence of word occurrences, regardless of other text components such as punctuation and space marks, as follows.

$$T = \{w_0, w_1, w_2, ..., w_{n-1}\} \tag{1}$$

Here, $w_i\ (0 \leq i \leq n-1)$ is the (i+1) th word occurrence of the text and $n$ is the length of the text. In Equation (1), we use the subscript of each word occurrence to identify its occurrence position. Furthermore, the subscripts are normalized with the word length of the text for the purpose of removing the influence of different text lengths. Then, the word occurrence position (or word position for short) of $w_i$ in text $T$ is defined as follows.

$$l_i = \frac{i}{n}, \text{ for } 0 \leq i \leq n-1 \tag{2}$$

Obviously, $0 \leq l_i < 1$. Therefore, Equations (1) and (2) represent all word occurrence information of the given text.

### B. NFZ Partition

In this step, all words in a given text are grouped into NFZs according to their NF values. Herein, the notions of NF and NFZ are defined as follows.

***Definition 1 Natural Frequency (NF)** of a word is the general frequency of the word in natural language texts, which can be evaluated by calculating the frequency of the word in a sufficiently large corpus.*

***Definition 2 Natural Frequency Zone (NFZ)** is a word group whose member words have similar NF values, where the NF similarity can be customized in accordance with specific conditions (See NFZ partition step).*

In statistical NLP area, word frequencies are of great importance for different kinds of analysis. In our style analysis, we make use of a NF dictionary and assign words to different NFZs according to their NF values. A NF dictionary can be regarded as a NF function mapping words to their corresponding NF values. Note that, if a word is not in the NF dictionary, we assign 0 to its corresponding NF.

Given NF function $y = f(x)$, where $x$ is any word and $y$ is its corresponding NF value, the NF set of text $T$ denoted by $F$ can be obtained as follows.

$$F = \{f(w_0), f(w_1), f(w_2), ..., f(w_{n-1})\} \tag{3}$$

Suppose that the maximal NF value in the NF dictionary is $f_{max}$, we can formalize the NFZs of text $T$ as follows.

$$Z_k = \xi(T, f(x), f_{max}), \text{ where } k = 1, 2, ..., K(f_{max}) \tag{4}$$

Herein, $\xi$ represents the way of NFZ partition and $k$ is the NFZ number. We have designed three ways of NFZ partition: linear partition, radix partition and logarithm partition.

***Definition 3 Linear partition***: *In linear partition, NFZs are of equal NF size. Suppose that the NF size is denoted by $L$ which is an integer, we group words with NF values $0, 1, 2, ..., L-1$ to the first NFZ, $L, L+1, L+2, ..., 2L-1$ to the second one and so on, i.e., $kL \leq f(w) < (k+1)L$ or $k = \left\lfloor \frac{f(w)}{L} \right\rfloor$. The NFZs are formalized as*

$$Z_k = \{w \mid k = \left\lfloor \frac{f(w)}{L} \right\rfloor, w \in T\} \tag{5}$$

*Herein, $k = 0, 1, 2, ..., K-1$, $K = \left\lfloor \frac{f_{max}}{L} \right\rfloor + 1$ is the count of NFZs.*

***Definition 4 Radix partition***: *In the radix partition, the NF sizes of NFZs are varying according to NF. The NFZs are designed as follows: the NF size of the first $R$ NFZs (denoted by $Z_0, Z_1, Z_2, ..., Z_{R-1}$) are $L$, that of the second $R$ NFZs are $RL$, that of the third $R$ NFZs are $R^2L$ and so on. The NFZs can be formalized as*

$$Z_k = \{w \mid k = \begin{cases} B & B < R \\ (R-1) \cdot E + \lfloor B / R^E \rfloor & B \geq R \end{cases}, w \in T\} \tag{6}$$

*Here, the parameter $R$ is subject to that $R > 1$ and $R$ is an integer. We call $R$ as the base radix and $L$ as the base size of NFZs. The item $B = \left\lfloor \frac{f(w)}{L} \right\rfloor$ is called the base frequency and the item $E = \lfloor \log_R B \rfloor$ the maximal exponent.*

***Definition 5 Logarithm partition***: *In the logarithm partition, the NFZs are formalized as*

$$Z_k = \{w \mid k = \begin{cases} 0 & f(w) = 0 \\ \lfloor \log_r f(w) \rfloor & f(w) > 0 \end{cases}, w \in T\} \tag{7}$$

*Here, the parameter $r$ is logarithm base and is subject to $r > 1$ and $r$ is a real.*

The following example illustrates how to partition a text in NFZs.

**Question**: Given $T = \{w_0, w_1, w_2, w_3, w_4\}$, $F = \{0, 80, 10000, 200000, 3000000\}$, find the NFZs.

**Solution**: For linear partition, given $L=100$, we have $Z_0 = \{w_0, w_1\}$, $Z_{100} = \{w_2\}$, $Z_{2000} = \{w_3\}$, $Z_{30000} = \{w_4\}$.

For radix partition, given $L=100$, $R=100$, we have $Z_0 = \{w_0, w_1\}$, $Z_{100} = \{w_2\}$, $Z_{119} = \{w_3\}$, $Z_{201} = \{w_4\}$.

For logarithm partition, given $r = 1.1$, we have $Z_0 = \{w_0\}$, $Z_{45} = \{w_1\}$, $Z_{96} = \{w_2\}$, $Z_{128} = \{w_3\}$, $Z_{156} = \{w_4\}$.

### C. NFZ Representation

After having formalized NFZs, a text $T$ can be regarded as a sequence of NFZ words from all the NFZs. Suppose that the text $T$ contains NFZ words from NFZ $Z_k$ with $n_k$ times, we have

$$\sum_{k=0}^{K-1} n_k = n \tag{8}$$

In other hand, the word position set of NFZ $Z_k$ can be obtained by Equation (1) and (2) and denoted by

$$L(Z_k) = \{l_{k0}, l_{k1}, l_{k2}, ..., l_{k,n_k-1}\} \tag{9}$$

This equation is subject to $l_{k0} < l_{k1} < l_{k2} < ... < l_{k,n_k-1}$.

Let $Z$ denote the set of NFZs, $L$ denote the set of $L(Z_k)$. That is

$$Z = \{Z_k \mid 0 \leq k < K\} \tag{10}$$

$$L = \{L(Z_k) \mid 0 \leq k < K\} \tag{11}$$

We can represent the text $T$ in NFZ representation form

$$T = <Z, L> \tag{12}$$

*D. Text Style Computation*

Based on the text formalization in Equation (12), in this step we can compute word distribution feature vector indicating text styles making use of the word occurrence information.

In order to measure the distribution of word occurrences, we first define the distance of word occurrences $w_i$ and $w_j$ in the text $T$ as

$$d_{ij} = d_{ji} = |l_i - l_j| \tag{13}$$

Then, we define in NFZ $Z_k$ Occurrence Distance Expectation (ODE) and Occurrence Distance Variance (ODV) as the average and variance of distances of neighboring word occurrences respectively, which are denoted by $\alpha_k$ and $\gamma_k$, as

$$\alpha_k = \frac{1}{n_k+1} \sum_{i=0}^{n_k} d_{i,i-1}^{(k)} = \frac{1}{n_k+1}(1-0) = \frac{1}{n_k+1} \tag{14}$$

$$\gamma_k = \frac{1}{\alpha_k} \sqrt{\frac{1}{n_k+1} \sum_{i=0}^{n_k} (d_{i,i-1}^{(k)} - \alpha_k)^2} \tag{15}$$

Here, $d_{i,i-1}^{(k)} = |l_i^{(k)} - l_{i-1}^{(k)}|$ denotes the distance of $w_i^{(k)}$ and $w_{i-1}^{(k)}$, which are the $(i+1)$th and $i$th word occurrences of NFZ $Z_k$. The boundary conditions are defined as $d_{0,-1}^{(k)} = |l_0^{(k)} - l_{-1}^{(k)}| = |l_0^{(k)} - 0| = l_0^{(k)}$ and $d_{n_k,n_k-1}^{(k)} = |l_{n_k}^{(k)} - l_{n_k-1}^{(k)}| = |1 - l_{n_k-1}^{(k)}| = 1 - l_{n_k-1}^{(k)}$, that is, we regard the beginning of the text ($l_{-1}^{(k)}=0$) and the end of the text ($l_{n_k}^{(k)}=1$) as two special word occurrences for any NFZ and thus there are $n_k+1$ word occurrence distances for NFZ $Z_k$, which is why we use $n_k+1$ in stead of $n_k$ in Equation (14) and (15).

ODE and ODV of an NFZ have obvious meanings: ODE implies the frequency of the NFZ word in the text, while ODV depicts the distribution of occurrences of the NFZ word in the text. For example, in Fig. 1, the word occurrence distances of NFZ word A in Text-1 are $\frac{0}{14}, \frac{2}{14}, \frac{1}{14}, \frac{3}{14}, \frac{3}{14}, \frac{1}{14}, \frac{2}{14}, \frac{1}{14}, \frac{1}{14}$ and ODE and ODV values are

$$\alpha_A = \frac{1}{n_A+1} = \frac{1}{8+1} = 0.1111$$

$$\gamma_A = \frac{1}{\alpha_A}\sqrt{\frac{1}{n_A+1}\sum_{i=0}^{n_A}(d_{i,i-1}^{(A)}-\alpha_A)^2} = 9\sqrt{\frac{1}{9}[(\frac{0}{14}-\frac{1}{9})^2+...(\frac{1}{14}-\frac{1}{9})^2]}=1.1737$$

In the same way, we can calculate ODE and ODV values for both Text-1 and Text-2 as

$$\begin{cases} \text{Text-1}: \alpha_A = 0.1111, \gamma_A = 1.1737; \alpha_B = 0.1429, \gamma_B = 1.1019 \\ \text{Text-2}: \alpha_A = 0.1111, \gamma_A = 1.3553; \alpha_B = 0.1429, \gamma_B = 1.3093 \end{cases}$$

We can see that though both texts have the same ODE values, Text-2 has greater ODV values than Text-1, indicating the difference of styles.

Finally, we combine all the ODE and ODV values as style feature vector $\Gamma$ as follows.

$$\Gamma = \{(\alpha_k, \gamma_k) | 0 \le k < K\} \quad (16)$$

Therefore, the vector $\Gamma$ represents the word distribution information of a text. As the size of NFZ decreases, it can depict the text style information to an inch.

### III. AUTHORSHIP ATTRIBUTION SCHEMES

In this section, we propose a basic authorship attribution scheme to solve closed authorship attribution problems, in which the author of the given text must be in a known candidate author set. Based on the basic authorship attribution scheme, we also propose an open authorship attribution scheme for the solution of open authorship attribution problems, in which the author of the given text is not necessary in a known candidate author set.

#### A. Basic Authorship Attribution Scheme

Normally, authorship attribution refers to attributing one author in a candidate set to a given testing text whose author is known to be among the candidate set. The version of authorship attribution is called closed authorship attribution in the paper. In our basic attribution scheme, we aim to effectively solve the closed authorship attribution problems.

In the basic authorship attribution scheme, there are two main procedures employed: feature extraction and classification. According to the style analytic method described in Section II, the following steps are applied to extract the style feature vector for a text being processed.

**Step 1: Word Occurrence Computation**. The algorithm reads the given text $T$, parses it,

splits it into words and obtains the word list $T$ in the form of Equation (1). Then the algorithm computes the word position of each word in the given text using Equation (2). See Alg. 1 in Fig. 2.

**Step 2: NFZ Partition & Representation**. The algorithm loads the NF dictionary, retrieves the NF values of words in the text $T$ and groups the words to NFZs according to their NF values. In this step, we use the linear partition as an example to illustrate NFZ partition and representation. At the end of the step, we get $Z_k$ with the corresponding word position set $L(Z_k)$, for $k = 0, 1, ..., K$. See Alg. 2 in Fig. 2.

**Step 3: Text Style Computation.** Using each word position set $L(Z_k)$, the algorithm computes the distances between any two neighboring occurrences in each NFZ $Z_k$, namely $d_{i,i-1}^{(k)}$ $(i = 0, 1, ..., n_k)$. Then, the algorithm computes ODE and ODV values of each NFZ $Z_k$ according to Equation (14) and (15). See Alg. 3 in Fig. 2.

**Alg. 1** Word Occurrence Computation

*Splits text $T$ into words and gets the total word count $n$.*
*For all $w_i \in T$ do*

$$l_i \leftarrow \frac{i}{n}$$

*End for*

**Alg. 2** NFZ Partition & Representation

*Loads NF dictionary and retrieves NF $f_i$ of each word in text $T$ to get NF set $F$.*
$Z_k \leftarrow \phi$
$L(Z_k) \leftarrow \phi$
*For all $w_i \in T$ do*

$$k \leftarrow \left\lfloor \frac{f_i}{L} \right\rfloor$$

$Z_k \leftarrow Z_k \cup \{w_i\}$
$L(Z_k) \leftarrow L(Z_k) \cup \{l_i\}$
*End for*

**Alg. 3** Text Style Computation

*For all $Z_k \in Z$ do*

$$\alpha_k \leftarrow \frac{1}{n_k + 1}$$

$\gamma_k \leftarrow 0$
*For $i \leftarrow 0..n_k$ do*

$$d_{i,i-1}^{(k)} \leftarrow |l_i - l_{i-1}|$$

$$\gamma_k \leftarrow \gamma_k + (d_{i,i-1}^{(k)} - \alpha_k)^2$$

*End for*

$$\gamma_k \leftarrow \frac{1}{\alpha_k} \sqrt{\frac{\gamma_k}{n_k + 1}}$$

*End for*

Fig. 2 Text Style Feature Extraction Algorithms

Going through all the steps described above, the analyzer converts the given text $T$ to a style feature vector $\Gamma$ as formalized in Equation (16). The vector $\Gamma$ is then used as the exclusive basis for the text style classification.

In the classification procedure, as Support Vector Machine (SVM) classifier is able to work effectively even when several thousands of features are used [2], we employ an existent SVM classifier LIBSVM [8].

The basic authorship attribution scheme is shown as Fig. 3. The arrowhead represents the flow of data. The dashed line arrowhead represents the training process that can be omitted if the classification model has already been prepared. The thick dashed rectangle indicates the whole analytic system.

Note that in the basic attribution scheme, we require that all training and testing texts are text segments roughly of the same word length. This requirement would avoid unbalance problem of text length and would make the analysis be more effective. The attribution results include the detailed information indicating which candidate author is attributed to each testing text, which will be used in the open authorship attribution scheme.

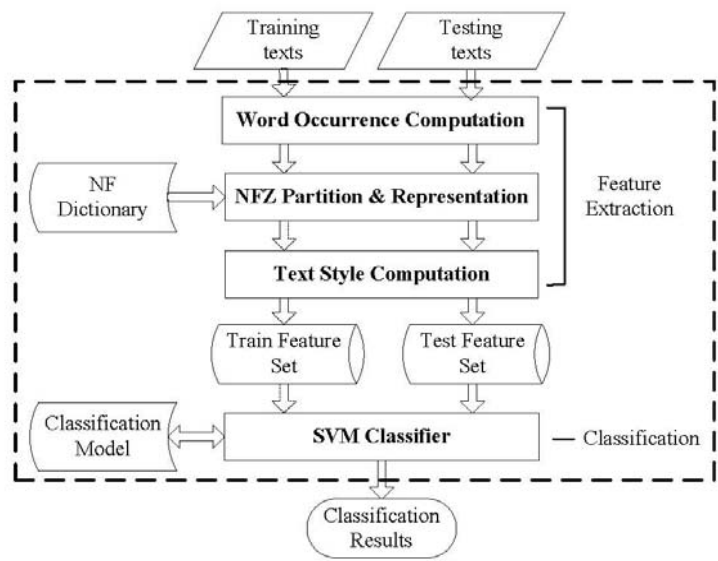

Fig. 3 the Basic Authorship Attribution Scheme of NFZ-WDA

## B. Open Authorship Attribution Scheme

The open authorship attribution refers to attributing one author in a candidate author set to a given text if its true author is among the candidate set, or rejecting all candidate authors as the author of the text if its true author is not among the candidate set.

Open authorship attribution problems are rarely addressed in the previous researches. They are obviously more difficult than closed attribution problems. In this section, we present an open attribution scheme making use of comparatively more sample texts by each author than the case of closed authorship attribution. More specifically, we assume that each testing text is long enough so that it can be split into some text segments to form a subset of sample texts.

Formally, we define the open authorship attribution problem as follows: given the sample text set $X=\{T_0, T_1, ..., T_{n-1}, T_n, T_{n+1}, ..., T_{n+m-2}, T_{n+m-1}\} \subset \mathcal{T}$, where set $T_i = \{t_{i,0}, t_{i,1}, ..., t_{i,n_i-1}\}$ is the sample text subset by author $i$, given the candidate author set $Y = \{a_0, a_1, ..., a_{n-2}, a_{n-1}\} \subset \mathcal{A}$, where element $a_i$ represents the author $i$, find a function $y = attr(T)$ satisfying

$$y = attr(T) = \begin{cases} a, & if\ auth(T) \in Y \\ \text{Reject}, & if\ auth(T) \notin Y \end{cases} \qquad (17)$$

Where $T \in X$, $a \in Y$ and $a = auth(T)$ represents the true author of sample text subset $T$.

The main ideal of our solution for the open authorship attribution problem is that if the true author of a sample text subset is among the candidate author set, then a test of closed authorship attribution on the subset will produce a result attributing most of the sample texts in the subset to the true author in the candidate set; otherwise, the result will attribute the sample texts comparatively equally among the candidate authors.

In practice, as absolutely accurate authorship attribution is impossible, we define the confidence of attributing author $a$ in candidate author set $Y$ to sample text subset $T$ as Equation (18).

$$f = conf_Y(T, a) = \frac{p_T(a) - 1/|Y|}{1 - 1/|Y|}, a \in Y \qquad (18)$$

Where $p_T(a)$ is the proportion of sample texts in subset $T$ that are attributed to author $a$. Here, the proportion $p_T(a)(a \in Y)$ can be obtained by test the sample text subset $T$ using the basic authorship attribution scheme; item $1/|Y|$ is the expected proportion for each author in the candidate set when using random attribution. In Equation (18), if $p_T(a) > 1/|Y|$, then $f > 0$, which indicates that the attribution to author $a$ is more possible than random attribution; otherwise, if $p_T(a) \leq 1/|Y|$, then $f \leq 0$, which indicates that the attribution is no more possible than the random attribution. Therefore, the greater the confidence $f$ is, the more possible the author $a$ is attributed to the sample text subset $T$.

***Definition 6*** *Sample text subset $T$ is $\theta$-**distinguishable** ($0 < \theta \leq 1$) on candidate author set $Y$ when*

1) $a = auth(T) \in Y$, if and only if $conf_Y(T, a) \geq \theta$ and $conf_Y(T, b) < \theta$ for all $b \in Y\text{-}\{a\}$
2) $a = auth(T) \notin Y$, if and only if $conf_Y(T, b) < \theta$ for all $b \in Y$

As a result, suppose that we can always find a proper $\theta (0 < \theta \leq 1)$ and make all the sample text subsets are $\theta$-distinguishable, the open authorship attribution scheme can be designed as follows.

**Step 1** For a given long sample text $t$, segment it into $n$ short sample texts $t_0, t_1, ..., t_{n-1}$, composing a sample text subset $T = \{t_0, t_1, ..., t_{n-1}\}$ with $|T| = n$.

**Step 2** Input the sample text subset $T = \{t_0, t_1, ..., t_{n-1}\}$ to the basic authorship attribution scheme, getting the proportion of sample texts that are attributed to each candidate author in the candidate author set $Y$, using Equation (18) to compute the confidence for each author.

**Step 3** Search the confidence values for all the candidate authors, if there is only one author with confidence not smaller than $\theta$, attribute the author to the sample text $t$; otherwise, reject to attribute any author to the sample text $t$.

## IV. EXPERIMENTS AND ANALYSIS

In this section, we experimentally apply both basic authorship attribution scheme and open authorship attribution scheme to analyzing texts written by different authors. We have designed four groups of experiments. The first group illustrates that the features employed in the proposed method are more than normal word frequency features. Next, the second one compares the effectiveness of the basic attribution scheme with that of Delta under various conditions. Then, the basic attribution scheme is tested with NF dictionaries generated from different corpuses and sample texts of blogs. Finally, the open attribution scheme is tested and compared to Delta.

In our experiments, we use 10 candidate authors with two prose works by each. Tab. 1 shows both training and testing data of the candidate author set and the word length distribution of the chapters. Because natural languages evolve as time goes, and so does the writing style of an author, we select prose works that are written in the same period between 1900 and 1920. We use one prose works to train and the other to test for each candidate author. For each prose works, we select 30 longest chapters (If there are less than 30 chapters in the prose works then use all the chapters.) as the training or testing set. In the experiments, both training and testing sample texts are in fact the front segments of the chapters, which are of a certain word length determined by a word length parameter, or the whole chapters if they are shorter than the designated length. For convenience, we combine the author's name, the writing year and the prose works' title to name the data set as shown in Tab. 1.

Tab. 1. Word Length Distribution of Chapters in Training and Testing Data of the Candidate Author Set

| NO | Chapter Sets of Prose Works | Number of Chapters of Different Word Length | | | | | | | | |
|----|----|----|----|----|----|----|----|----|----|----|
| | | <=1000 | 10−1500 | 15−2000 | 20−2500 | 25−3000 | 30−4000 | 40−5000 | >5000 | Total |
| A0 | Anderson.1917.Marching-Men.train | 0 | 6 | 7 | 7 | 4 | 3 | 3 | 0 | 30 |
| | Anderson.1919.Winesburg-Ohio.test | 0 | 5 | 1 | 4 | 4 | 8 | 1 | 1 | 24 |
| A1 | Chesnutt.1900.The-House-Behind-the-Cedars.train | 2 | 5 | 10 | 6 | 4 | 1 | 1 | 1 | 30 |
| | Chesnutt.1901.The-Marrow-of-Tradition.test | 0 | 0 | 8 | 8 | 1 | 11 | 2 | 0 | 30 |
| A2 | Dixon.1905.The-Clansman.train | 0 | 1 | 6 | 10 | 4 | 6 | 3 | 0 | 30 |
| | Dixon.1902.The-Leopards-Spots.test | 0 | 0 | 0 | 7 | 11 | 10 | 1 | 1 | 30 |
| A3 | Dreiser.1900.Sister-Carrie.train | 0 | 0 | 0 | 0 | 3 | 19 | 6 | 2 | 30 |
| | Dreiser.1914.The-Titan.test | 0 | 0 | 0 | 0 | 0 | 16 | 14 | 0 | 30 |
| A4 | Glasgow.1900.The-Voice-of-the-People.train | 0 | 0 | 0 | 0 | 10 | 20 | 0 | 0 | 30 |
| | Glasgow.1902.The-Battle-Ground.test | 0 | 0 | 0 | 0 | 10 | 15 | 4 | 1 | 30 |
| A5 | James.1909.The-Ambassadors.train | 0 | 0 | 0 | 0 | 0 | 12 | 10 | 8 | 30 |
| | James.1909.The-Wings-of-Dove.test | 0 | 0 | 0 | 0 | 0 | 2 | 12 | 16 | 30 |
| A6 | London.1910.Burning-Daylight.train | 0 | 0 | 0 | 9 | 10 | 6 | 5 | 0 | 30 |
| | London.1906.White-Fang.test | 0 | 0 | 7 | 3 | 5 | 9 | 1 | 0 | 25 |
| A7 | Philips.1917.Susan-Lenox.train | 0 | 0 | 0 | 0 | 0 | 0 | 0 | 29 | 29 |
| | Philips.1911.The-Dust.test | 0 | 0 | 1 | 0 | 0 | 8 | 8 | 5 | 22 |
| A8 | Sinclair.1906.The-Jungle.train | 0 | 0 | 0 | 0 | 0 | 7 | 13 | 10 | 30 |
| | Sinclair.1908.The-Metropolis.test | 0 | 0 | 0 | 0 | 0 | 2 | 11 | 8 | 21 |
| A9 | Stratton-Porter.1904.A-Girl-of-the-Limberlost.train | 0 | 0 | 0 | 1 | 1 | 8 | 7 | 8 | 25 |
| | Stratton-Porter.1904.Freckles.test | 0 | 0 | 0 | 4 | 1 | 6 | 5 | 4 | 20 |

In the experiments, if no specific instruction, the configures are as follows: NF dictionary is the word frequency list in descending order obtained from BNC corpus [9]; the first hundreds of words in NF dictionary are also used in Delta analysis as the most frequent words; the NFZ partition parameters are set by experience as follow: for linear partition, $L=10$; for radix partition, $L=10$, $R=100000$; for logarithm partition, $r=1.0001$.

### A. More than Word Frequencies

In NFZ-WDA style analysis, style feature set can be divided into two parts: ODE subset and ODV subset. In order to evaluate the effectiveness of each part of style feature set, we compare the analysis using the whole feature set with the analysis using only one feature subset.

Fig. 4 depicts the attribution accuracies for sample texts of different word lengths over the NFZ-WDA analysis with linear partition using the whole feature set, that using only ODE subset and that using only ODV subset. According to the definition of ODE, ODE features are in fact equivalent to word frequency features of NFZ words. As a result, authorship attribution with only ODE features can indicate the effect of word frequencies on distinguishing the styles of different authors.

In Fig. 4, we can see that attribution using the whole feature set provides steadily better results than that using only one feature subset, while the attribution using only ODE subset provides better results than that using ODV subset. Thus, the experimental results confirm that the distribution features of NFZ word occurrences provide more style details than normal word frequencies, which also make the NFZ-WDA be more effective than normal word frequency analysis. In fact, even the ODE subset is different from the previous word frequency analysis such as in Delta, since in NFZ-WDA all words are grouped into different NFZs according merely to their NF values and all words contribute to the text style analysis.

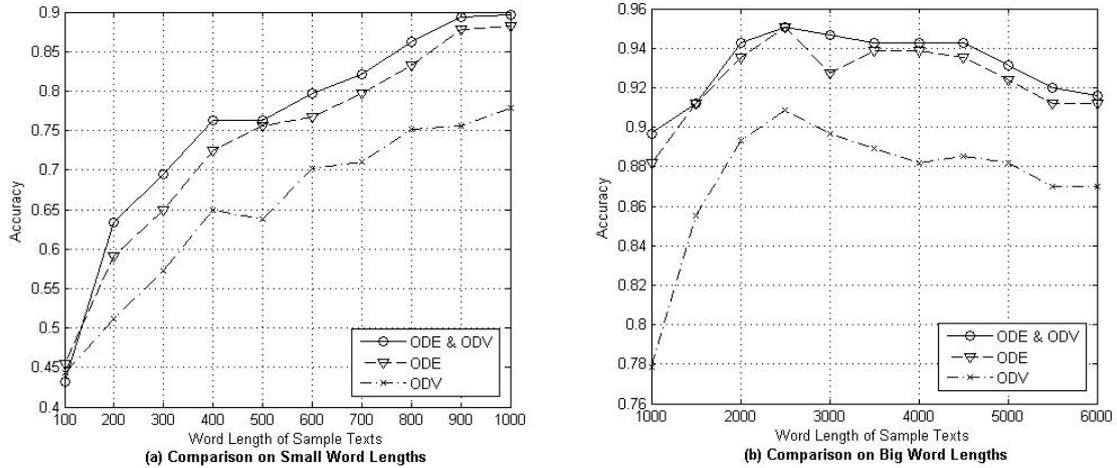

Fig. 4 The Comparison of NFZ-WDA When using Different Parts of Style Features

### B. Comparison between NFZ-WDA and Delta

Delta is one of the most promising developments in authorship attribution in recent years, which has been first introduced by John F. Burrows in his Busa Award presentation [10]. Later, Delta is further tested and developed by Hoover [3][11]. In our experiments, we implement the Delta method using the most frequent words obtained from British National Corpus (BNC) [9], and compare the effectiveness of NFZ-WDA with that of Delta analysis.

Fig. 5 illustrates the comparisons between NFZ-WDA and Delta analysis under various conditions. Fig. 5(a) depicts the comparison between NFZ-WDA (radix partition) and Delta (150 most frequent words) on different numbers of candidate authors. We use sample texts of 1000 word length and the training and testing data as shown in Tab. 1. The author number changes from 2 to 10: for 2 to 6, we randomly select the numbers of authors from the candidate author set and repeat 30 times for averaging; for 7 to 10, we do the same thing except repeating 10 times. Fig. 5(a) shows that NFZ-WDA has much better results than Delta for all numbers of authors while the accuracies of both fall as the author number increases.

Fig. 5(b) shows the comparison also between NFZ-WDA (radix partition) and Delta (150 most frequent words) using different numbers of training and testing texts. The X values $1, 2, ..., 6, 7, 8, ..., 35, 36$ represents the number pair of training and testing texts

$(5,5),(5,10),...,(5,30),(10,5),(10,10),...,(30,25),(30,30)$. We can see that NFZ-WDA has better results than Delta in all cases of training and testing text configures while the accuracies of both increase as the number of training texts increases.

As depicted in Fig. 5(c) and (d), the attribution accuracies for sample texts of different word lengths over NFZ-WDA with linear partition, radix partition and logarithm partition are compared with the attribution accuracies or top 2 or 3 attribution accuracies over Delta methods with 150 and 700 most frequent words. The reason for selecting 150 and 700 most frequent words is that the test with 150 words can roughly simulate the original test of Burrows with 150 most frequent words and the test with 700 words is considered to be the most effective [11]. In Fig. 5(c) and (d), it turns out that: 1) NFZ-WDA methods with different NFZ partitions have much better results than the Delta methods with different numbers of most frequent words, and are even comparative to the top 2 or top 3 attributions with 150 most frequent words; 2) In the three NFZ partitions, logarithm partition produces the best results.

In short, the experimental results depicted in Fig. 5 show NFZ-WDA outperforms Delta analysis in solutions for closed authorship attribution problems on the same training and testing data. The reasons are probably as follows: 1) NFZ-WDA analysis makes use of all the words contained in the sample texts while Delta analysis makes use of only hundreds of the most frequent words, by which the former can capture more style details. 2) The use of word occurrence level information provides NFZ-WDA analysis even more style details. 3) The application of SVM classifier makes NFZ-WDA analysis more effective.

Furthermore, from Fig. 5(c) and (d), we can see that NFZ-WDA with logarithm partition performs the best. The reasons may be that logarithm partition provides more reasonable NFZ partitions. In natural language texts, word types become sparser as their NF values increase, so an NFZ partition that makes the NFZs with greater NF values be of larger sizes seems to be more effective for the attribution. From the NFZ partition example in Section II, we can see that logarithm partition satisfies the requirements better.

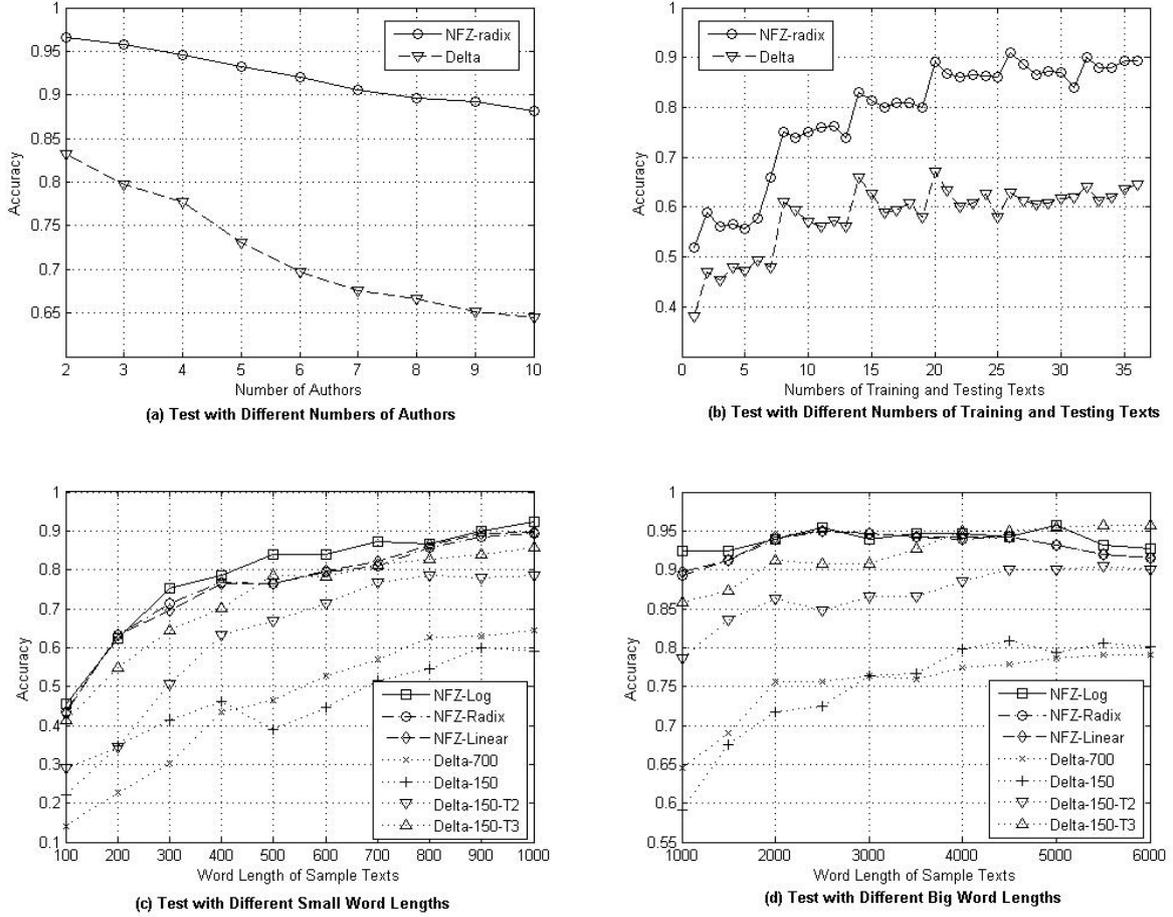

Fig. 5 The Comparisons between NFZ-WDA and Delta under Various Conditions

## C. Tests with Other NF Dictionaries and Sample Texts

In this group of experiments, we test the basic scheme with NF dictionaries generated from different corpora and with blog texts.

We use three different NF dictionaries: NF dictionary from BNC Corpus, NF dictionary from the Corpus of Contemporary American English (COCA) [12] and NF dictionary from the literature works gathered by us (LIT). We use radix partition with $L=10$, $R=100000$ for all the three cases. The word length of sample texts is 1000 and the training and testing texts are shown as Tab. 1. The experimental results depicted in Fig. 6(a) show roughly the same trend line for the three cases, which indicates that NFZ-WDA with different NF dictionaries generated from different corpora performs similarly.

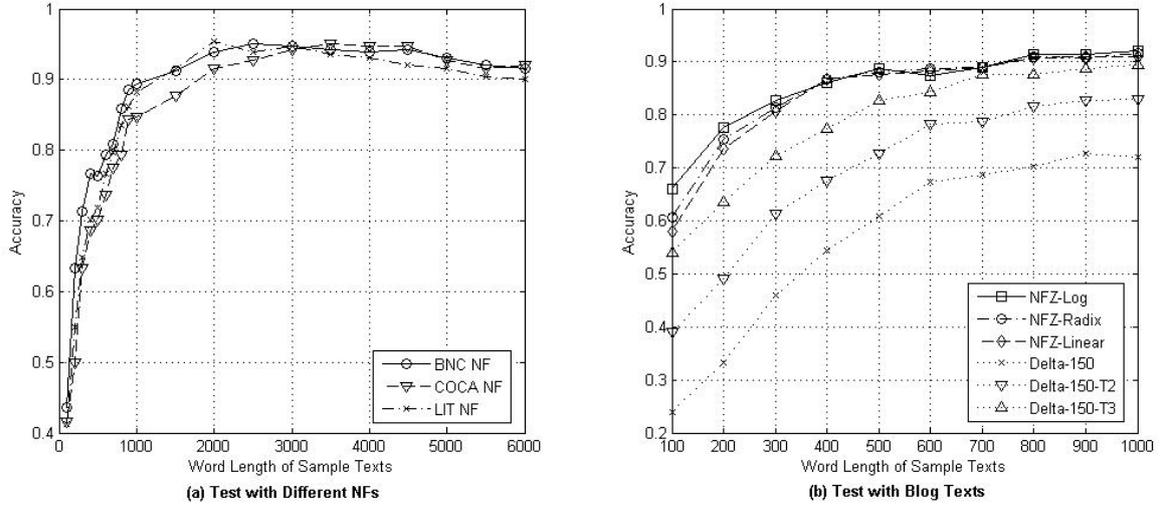

Fig. 6 Tests with Other NF Dictionaries and Sample Texts

For further verification, we test NFZ-WDA and Delta with blog texts extracted from the blog corpus [13].The blog texts of 10 authors as shown in Tab. 2 are used. 80 blog texts of lengths more than 3000 characters (about 500 words) are selected for each author (with 50 texts as training texts and 30 texts as testing texts). As the lengths of most blog texts are between 500 and 1000 words, we make the word length change from 100 to 1000. Fig. 6(b) shows the experimental results on the blog texts. Compared to Fig. 5(c), both NFZ-WDA and Delta perform better on blog texts than on prose texts and still NFZ-WDA outperforms Delta. It seems that authorship attribution for blog texts is easier than texts of prose works of the same length, for the writing time for blogs by the same author is often very close and the writing style changes little.

Tab. 2 The 10 Blog Sets Selected

| No | Blog Text Set | No | Blog Text Set |
|---|---|---|---|
| 1 | 1516660.male.17.Student.Cancer | 6 | 1151815.male.25.Education.Leo |
| 2 | 1476382.male.33.Publishing.Gemini | 7 | 1107146.female.16.Student.Libra |
| 3 | 1417798.female.35.indUnk.Scorpio | 8 | 1046946.female.25.Arts.Virgo |
| 4 | 1234212.male.27.indUnk.Taurus | 9 | 216413.female.26.Law.Cancer |
| 5 | 1169141.male.33.Arts.Capricorn | 10 | 122217.male.37.indUnk.Leo |

## D. Open Authorship Attribution

In this group of experiments, we first show that by using our basic authorship attribution scheme the sample texts by authors in the candidate author set can be attributed to the true author with an especially high possibility, while those by authors out of the candidate author set are attributed comparatively equally among the candidate authors. As depicted in Tab. 3, the sample texts of 400 word length by authors A0-A5 are attributed to themselves with high possibilities of 0.8750, 0.7000, 0.7667, 0.6333, 0.8667and 0.7667respectively, which is far greater than the possibilities of being attributed to other authors; while the sample texts by authors A6*, A7*, A8* and A9* out of the candidate author set are attributed more equally among the candidate authors (with the greatest attribution possibilities being 0.4000, 0.3333, 0.2727 and 0.2500 respectively).

Tab. 3 The Attribution Results When some Sample Text Subsets are written by Authors out of the Candidate Author Set with Word Length being 400

| Authorship Attribution | | Attributed Authors | | | | | | | | | |
|---|---|---|---|---|---|---|---|---|---|---|---|
| | | A 0 | A 1 | A 2 | A 3 | A 4 | A 5 | A 6 | A 7 | A 8 | A 9 |
| True Authors | A 0 | **0.8750** | 0.1250 | 0.0000 | 0.0000 | 0.0000 | 0.0000 | 0.0000 | 0.0000 | 0.0000 | 0.0000 |
| | A 1 | 0.0000 | **0.7000** | 0.0000 | 0.0000 | 0.1333 | 0.0000 | 0.0333 | 0.1000 | 0.0000 | 0.0333 |
| | A 2 | 0.0000 | 0.0000 | **0.7667** | 0.0000 | 0.0333 | 0.0000 | 0.0000 | 0.0333 | 0.0000 | 0.1667 |
| | A 3 | 0.0000 | 0.0667 | 0.0667 | **0.6333** | 0.0000 | 0.0000 | 0.0000 | 0.1333 | 0.1000 | 0.0000 |
| | A 4 | 0.0000 | 0.1000 | 0.0000 | 0.0000 | **0.8667** | 0.0000 | 0.0000 | 0.0333 | 0.0000 | 0.0000 |
| | A 5 | 0.0000 | 0.0000 | 0.0000 | 0.1333 | 0.0000 | **0.7667** | 0.0000 | 0.1000 | 0.0000 | 0.0000 |
| | A 6* | 0.0000 | 0.2667 | 0.0000 | 0.2000 | 0.0667 | 0.0000 | 0.0667 | **0.4000** | 0.0000 | 0.0000 |
| | A 7* | 0.0000 | 0.0000 | 0.2222 | **0.3333** | 0.2222 | 0.0000 | 0.0000 | 0.2222 | 0.0000 | 0.0000 |
| | A 8* | 0.0455 | 0.0455 | 0.0000 | 0.2273 | **0.2727** | 0.0000 | 0.0000 | 0.2273 | 0.0455 | 0.1364 |
| | A 9* | 0.0000 | 0.1429 | 0.1786 | 0.1786 | 0.1071 | 0.0000 | 0.0357 | **0.2500** | 0.1071 | 0.0000 |

A6* =Cabell.1919.Jurgen-A-Comedy-of-Justice.test (30 chapters)  A7* = Fitzgerald.1920.This-Side-of-Paradise.test (9 chapters)
A8* = Baum.1900.The-Wonderful-Wizard-of-Oz.test (22 chapters)  A9* = Burroughs.1914.Tarzan-of-the-Apes.test (28 chapters)

Then, we test our open authorship attribution method on 35 prose works with 20 ones by the authors out of and 15 ones in the candidate author set. Fig. 7(a) and (b) show the maximum confidence for each prose works, when word length = 400, confidence threshold = 0.5 and word length = 1000, confidence threshold = 0.5. Tab. 4 and Tab. 5 show the detailed attribution results of the 35 prose works when word length = 400, confidence threshold = 0.5.

From Fig. 7(a) and (b), we can see that the attribution accuracies are the same, 97.14% (=34/35), for both cases. So, the word length 400 seems to be enough for open attribution test, while the increase of the word length only makes the confidence differences of sample texts written by authors in or out of candidate set be more obvious. Furthermore, in both Fig. 7(a) and (b), there is still one prose works by author in the candidate author set is wrongly rejected. It seems that increasing the word length makes the prose more tend to be accepted. Referring to Tab. 5, we can see that the wrongly rejected prose works is James' prose entitled "The Europeans" written in 1878. The rejecting of the prose "The Europeans" seems to imply that James' style changes a lot from 1878 to 1909 (Note that the prose "The ambassadors" written in 1909 is used as the training set).

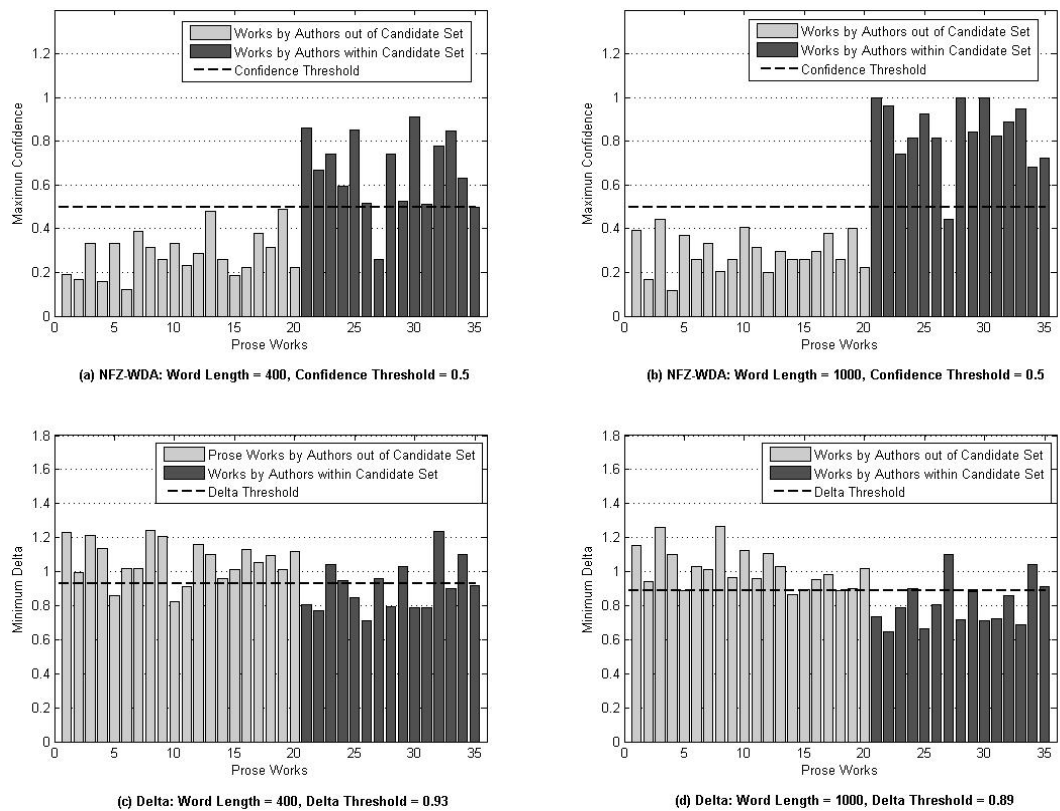

Fig. 7 Open Authorship Attribution Results for the 35 Prose Works using Both NFZ-WDA and Delta

Fig. 7(c) and (d) show the experimental results of test Delta using the same sample texts and the same word lengths. We can see that about 9 prose works are wrongly accepted or rejected in the case of word length = 400, Delta threshold = 0.93, while about 5 ones in the case of word length = 1000, Delta threshold = 0.89. The results are much worse than those reported in the paper [10], since our experiments only use part of each prose works. For example, if word length = 400, the number of the chapters of a prose is 10, then we at most use (400×10 = ) 4000 words of texts of the prose.

It is obvious that in the experiments our open attribution scheme produces better results than Delta. Furthermore, the confidence threshold values of our scheme is more meaningful and easier to be decide on (can simply use the threshold value 0.5 for different word lengths) than the Delta threshold values of Delta, which are hard to be properly decided on if the true authors of the tested works are unknown.

Tab. 4. Detailed Open Attribution Results for the 20 Prose Works out of the Candidate Author Set
(NFZ-WDA: Word Length = 400, Confidence Threshold = 0.5; A/R: Accepted/Rejected; TA: True Author; AA: Attributed Author)

| No | Prose Works | A/R | TA | AA | MaxConf | A0 | A1 | A2 | A3 | A4 | A5 | A6 | A7 | A8 | A9 | Tol |
|---|---|---|---|---|---|---|---|---|---|---|---|---|---|---|---|---|
| 1 | Baum.1900.The-Wonderful-Wizard-of-Oz | R | NA | -- | 0.1919 | 1 | 1 | 0 | 5 | **6** | 0 | 0 | 5 | 1 | 3 | 22 |
| 2 | Burroughs.1914.Tarzan-of-the-Apes | R | NA | -- | 0.1667 | 0 | 4 | 5 | 5 | 3 | 0 | 1 | **7** | 3 | 0 | 28 |
| 3 | Cabell.1919.Jurgen-A-Comedy-of-Justice | R | NA | -- | 0.3333 | 0 | 8 | 0 | 6 | 2 | 0 | 2 | **12** | 0 | 0 | 30 |
| 4 | Cather.1925.The-Professors-House | R | NA | -- | 0.1571 | 1 | 4 | 0 | 3 | 3 | 0 | **7** | **7** | 2 | 2 | 29 |
| 5 | Chopin.1899.The-Awakening | R | NA | -- | 0.3333 | 0 | **12** | 0 | 2 | 3 | 1 | 0 | 11 | 0 | 1 | 30 |
| 6 | Crane.1895.The-Red-Badge-of-Courage | R | NA | -- | 0.1204 | 1 | 0 | 2 | 4 | 4 | 1 | 3 | 4 | **5** | 0 | 24 |
| 7 | Davis.1897.Frances-Waldeaux | R | NA | -- | 0.3889 | 0 | 0 | 0 | 3 | 1 | 0 | 0 | **9** | 1 | 6 | 20 |
| 8 | Faulkner.1932.Light-in-August | R | NA | -- | 0.3122 | 0 | 0 | 0 | 5 | 2 | 0 | 0 | 3 | **8** | 3 | 21 |
| 9 | Fitzgerald.1920.This-Side-of-Paradise | R | NA | -- | 0.2593 | 0 | 0 | 2 | **3** | 2 | 0 | 0 | 2 | 0 | 0 | 9 |
| 10 | Frederic.1896.The-Damnation-of-Theron-Ware | R | NA | -- | 0.3333 | 0 | 3 | 0 | **12** | 0 | 0 | 2 | 8 | 5 | 0 | 30 |
| 11 | Grey.1919.The-Man-of-the-Forest | R | NA | -- | 0.2308 | 0 | 1 | 3 | **8** | 3 | 0 | 1 | 7 | 1 | 2 | 26 |
| 12 | Howells.1890.The-Kentons | R | NA | -- | 0.2889 | 0 | **9** | 0 | 4 | 0 | 5 | 0 | 3 | 0 | 4 | 25 |
| 13 | Jewett.1884.A-Country-Doctor | R | NA | -- | 0.4815 | 1 | **16** | 0 | 4 | 3 | 2 | 1 | 3 | 0 | 0 | 30 |
| 14 | Lewis.1922.Babbitt | R | NA | -- | 0.2593 | 0 | 1 | 7 | **10** | 3 | 0 | 2 | 4 | 1 | 2 | 30 |
| 15 | Norris.1901.The-Octopus | R | NA | -- | 0.1852 | 0 | 1 | **4** | 1 | 2 | 0 | 1 | 2 | 3 | 1 | 15 |
| 16 | Rinehart.1919.Dangerous-Days | R | NA | -- | 0.2222 | 0 | 0 | 3 | 8 | 1 | 1 | 0 | **9** | 2 | 6 | 30 |
| 17 | Tarkington.1921.Alice-Adams | R | NA | -- | 0.3778 | 0 | 0 | 0 | 4 | 0 | 1 | 0 | **11** | 1 | 8 | 25 |

| NO | Prose Works | A/R | TA | AA | MaxConf | A0 | A1 | A2 | A3 | A4 | A5 | A6 | A7 | A8 | A9 | Tol |
|---|---|---|---|---|---|---|---|---|---|---|---|---|---|---|---|---|
| 18 | Twain.1894.The-Tragedy-of-Pudd-Nhead-Wilson | R | NA | -- | 0.3122 | 0 | 1 | 1 | 2 | 2 | 0 | 0 | 6 | **8** | 1 | 21 |
| 19 | Wharton.1905.The-House-of-Mirth | R | NA | -- | 0.4872 | 0 | 3 | 0 | 0 | 2 | 1 | 0 | **7** | 0 | 0 | 13 |
| 20 | White.1904.The-Silent-Places | R | NA | -- | 0.2222 | 0 | 0 | **9** | 6 | 1 | 0 | 1 | 8 | 3 | 2 | 30 |
| Rates | | Accepted: 0%(0/20); Rejected: 100%(20/20) | | | | | | | | | | | | | | |

Tab. 5. Detailed Open Attribution Results for the 15 Prose Works in the Candidate Author Set

(NFZ-WDA: Word Length = 400, Confidence Threshold = 0.5; A/R: Accepted/Rejected; TA: True Author; AA : Attributed Author)

| NO | Prose Works | A/R | TA | AA | MaxConf | A0 | A1 | A2 | A3 | A4 | A5 | A6 | A7 | A8 | A9 | Tol |
|---|---|---|---|---|---|---|---|---|---|---|---|---|---|---|---|---|
| 21 | Anderson.1919.Winesburg-Ohio | A | A0 | A0 | 0.8611 | **21** | 3 | 0 | 0 | 0 | 0 | 0 | 0 | 0 | 0 | 24 |
| 22 | Chesnutt.1901.The-Marrow-of-Tradition | A | A1 | A1 | 0.6667 | 0 | **21** | 0 | 0 | 4 | 0 | 1 | 3 | 0 | 1 | 30 |
| 23 | Dixon.1902.The-Leopards-Spots | A | A2 | A2 | 0.7407 | 0 | 0 | **23** | 0 | 1 | 0 | 0 | 1 | 0 | 5 | 30 |
| 24 | Dreiser.1914.The-Titan | A | A3 | A3 | 0.5926 | 0 | 2 | 2 | **19** | 0 | 0 | 0 | 4 | 3 | 0 | 30 |
| 25 | Glasgow.1902.The-Battle-Ground | A | A4 | A4 | 0.8519 | 0 | 3 | 0 | 0 | **26** | 0 | 0 | 1 | 0 | 0 | 30 |
| 26 | Glasgow.1904.The-Deliverance | A | A4 | A4 | 0.5185 | 0 | 0 | 2 | 3 | **17** | 0 | 0 | 6 | 0 | 2 | 30 |
| 27 | James.1878.The-Europeans | R | A5 | -- | 0.2593 | 0 | 2 | 0 | **4** | 2 | 1 | 0 | 3 | 0 | 0 | 12 |
| 28 | James.1909.The-Wings-of-Dove | A | A5 | A5 | 0.7407 | 0 | 0 | 0 | 4 | 0 | **23** | 0 | 3 | 0 | 0 | 30 |
| 29 | London.1903.The-Call-of-the-Wild | A | A6 | A6 | 0.5238 | 0 | 0 | 0 | 1 | 0 | 0 | **4** | 0 | 2 | 0 | 7 |
| 30 | London.1906.White-Fang | A | A6 | A6 | 0.9111 | 0 | 0 | 0 | 0 | 0 | 0 | **23** | 1 | 1 | 0 | 25 |
| 31 | Philips.1909.The-Fashionable-Adventures | A | A7 | A7 | 0.5111 | 0 | 0 | 1 | 7 | 1 | 0 | 0 | **14** | 1 | 1 | 25 |
| 32 | Philips.1911.The-Conflict | A | A7 | A7 | 0.7778 | 0 | 2 | 0 | 0 | 0 | 0 | 0 | **8** | 0 | 0 | 10 |
| 33 | Philips.1911.The-Dust | A | A7 | A7 | 0.8485 | 0 | 1 | 0 | 2 | 0 | 0 | 0 | **19** | 0 | 0 | 22 |
| 34 | Sinclair.1908.The-Metropolis | A | A8 | A8 | 0.6296 | 0 | 0 | 2 | 2 | 0 | 0 | 0 | 1 | **14** | 2 | 21 |
| 35 | Stratton-Porter.1904.Freckles | A | A9 | A9 | 0.5000 | 0 | 0 | 2 | 1 | 0 | 0 | 2 | 2 | 2 | **11** | 20 |
| Rates | | Accepted: 93.33%(14/15); Rejected: 6.67%(1/15); Attribution Accuracy: 100%(14/14) | | | | | | | | | | | | | | |

## V. RELATED WORK

From a machine learning perspective, the two most important factors of authorship attribution involve the choice of features and the selection of an appropriate and effective classification technique [1].

In the previous research, a great number of textual features for authorship attribution have been proposed, including lexical features, character features, syntactic features, semantic features and application-specific features [2]. Lexical features can be defined, from the simple measures such as sentence length and word length [14] to the vocabulary richness features, and then to the vectors of function word frequencies and even word n-gram frequencies. Character features includes alphabetic character count, digit character count, uppercase and lowercase character count, letter frequencies, punctuation marks count, character n-gram frequencies, and so on [15][16]. Syntactic features refer to syntactic patterns extracted by NLP tools and they require robust and accurate NLP tools able to perform syntactic analysis of texts. Very few attempts have been made to exploit semantic features for authorship attribution purposes because of the difficulties of complicated tasks of text analysis. Besides, application-specific features can better represent the nuances of style in a given text domain. For example, in domains such as e-mail massages and online forum messages, structural measures, including the use of greetings and farewells, types of signatures, etc [4][14][17][18], can be used.

Among all these features, the most frequent words, or the function words are found to be the best features to discriminate between authors [19][20]. The most notable method of using the most frequent words has been proposed by Burrows [10] under the name 'Delta'. Delta works as follows. First, the method calculates the z distributions of a set of function words (originally, the 150 most frequent words). Then, for each text, the deviation of each word frequency from the norm is calculated in terms of z-score, roughly indicating whether it is used more (positive z-score) or less (negative z-score) times than the average. Finally, the Delta measure is the mean of the absolute differences between the z-scores for the entire function word set in a set of training texts written by the same author and the corresponding z-scores of an unknown text, indicating the difference between the training texts and the unknown text. The smaller Delta measure the greater stylistic similarity between the unknown text and the candidate author.

Together with the features proposed above, classification methods such as support vector machine (SVM), k-nearest neighbors (KNN) and regularized discriminant analysis (RDA), can be applied to authorship attribution [1]. Among these methods, SVM is able to properly deal with the high dimension feature vector even with several thousands of features and is considered one of the best classifier [4].

VI. CONCLUSIONS AND FUTURE WORK

In this paper, we have further developed and tested the text style analysis called natural frequency zoned word distribution analysis (NFZ-WDA) that was first proposed in our conference paper [5]. NFZ-WDA introduces notions of NFZ partition and word occurrence information for text style analysis, based on the observation that all authors leave distinct intrinsic word usage style when writing. Using the style analysis, we have proposed a basic authorship attribution scheme and an open authorship attribution scheme, for solving both closed and open authorship attribution problems. Especially, the open attribution scheme provides a novel ideal for solution of open authorship attribution problems which are scarcely addressed in previous research. Extensive experiments have shown that NFZ-WDA is fairly promising in solution of authorship attribution problems.

The future work is to apply NFZ-WDA to authorship attribution for other natural languages and more kinds of practical texts under more strict conditions. Our current work has mainly focused on English and the literature, blog texts. Other natural languages and more kinds of texts, such as the texts of emails, newspapers, messages, should be considered. More practical conditions, such as the conditions of limited data and various genres, should also be taken into account.